\documentclass[runningheads]{llncs}
\usepackage[T1]{fontenc}
\usepackage{booktabs}
\usepackage{multirow}
\usepackage{adjustbox}
\usepackage{setspace}
\usepackage{xspace}
\usepackage{amssymb}
\usepackage{wasysym}
\usepackage{enumitem}
\usepackage{tikz}

\usepackage{hyperref}
\hypersetup{
    colorlinks=true, 
    linkcolor=blue,  
    filecolor=blue,  
    urlcolor=blue,   
    citecolor=blue,  
}

\usepackage[caption=false]{subfig}
\usepackage{overpic} 

\usepackage{graphicx}
\newcommand{\etal}{\textit{et al.}\xspace}

\newcommand{\our}{{PPAD}\xspace}
\newcommand{\chest}{{chest X-ray}\xspace}

\begin{document}
\title{Position-Guided Prompt Learning for Anomaly Detection in Chest X-Rays}

%
%
\author{Zhichao Sun\inst{1} \and
Yuliang Gu\inst{1} \and
Yepeng Liu\inst{1} \and
Zerui Zhang\inst{1} \and
Zhou Zhao\inst{2}  \and
Yongchao Xu\inst{1}
}
\authorrunning{Z. Sun et al.}
%
\institute{School of Computer Science, Wuhan University, Wuhan, China \newline
\email{\{zhichaosun, yongchao.xu\}@whu.edu.cn}\\ \and
School of Computer Science, Central China Normal University, Wuhan, China
\email{zhaozhou@ccnu.edu.cn}}

%
\maketitle              
\begin{abstract}
Anomaly detection in chest X-rays is a critical task. Most methods mainly model the distribution of normal images, and then regard significant deviation from normal distribution as anomaly. Recently, CLIP-based methods, pre-trained on a large number of medical images, have shown impressive performance on zero/few-shot downstream tasks. In this paper, we aim to explore the potential of CLIP-based methods for anomaly detection in chest X-rays. Considering the discrepancy between the CLIP pre-training data and the task-specific data, we propose a position-guided prompt learning method. Specifically, inspired by the fact that experts diagnose chest X-rays by carefully examining distinct lung regions, we propose learnable position-guided text and image prompts to adapt the task data to the frozen pre-trained CLIP-based model.
To enhance the model's discriminative capability, we propose a novel structure-preserving anomaly synthesis method within chest x-rays during the training process. Extensive experiments on three datasets demonstrate that our proposed method outperforms some state-of-the-art methods. The code of our implementation is available at \url{https://github.com/sunzc-sunny/PPAD}.
\end{abstract}
\section{Introduction}

Chest X-ray remains the most frequently utilized diagnostic tool for detecting lung diseases. Thus, the rise of automated chest X-ray analysis through deep learning techniques has gained considerable attention. However, pathology annotations of chest X-rays are hard to obtain, requiring expert experience. Consequently, unsupervised anomaly detection methods~\cite{zhou2020encoding,zhou2021proxy} are proposed, aiming to identify diseases in the testing set using only normal data from the training set. Previous methods~\cite{salad,if2d,squid,amae,baugh2023many,ddad} 
mainly attempt to model the distribution of normal images during training, and then identify the samples that do not conform to normal profile as anomalies. 

Recently, Constrastive Language-Image Pre-training (CLIP)~\cite{clip} has gained increasing attention in the field of medical imaging. Advancements~\cite{chexzero,you2023cxr,chen2023knowledge} demonstrate the effectiveness of CLIP-based methods in downstream tasks when pre-trained on large-scale chest X-ray datasets with image-text pairs. Specifically, CheXzero~\cite{chexzero} pre-trains CLIP on a large chest X-ray dataset for zero-shot classification. KoBo~\cite{chen2023knowledge} integrates clinical knowledge into the pre-training process, achieving comparable results in eight downstream tasks. Despite the effectiveness of CLIP for downstream tasks, its massive parameters and extensive training data requirement make it impractical to fine-tune the full model for specific tasks. To address this challenge, prompt engineering and prompt learning methods~\cite{xplainer,zhou2022learning,khattak2023maple} have been proposed. For instance, Xplainer~\cite{xplainer} focuses on designing appropriate expert-level text input in prompt engineering. CoOp~\cite{zhou2022learning} proposes learnable text contexts for prompt learning. MaPLe~\cite{khattak2023maple} presents a multi-modal prompt learning framework for both vision and language branches of CLIP. However, these methods do not directly consider the domain differences between pre-training data and task-specific data. Besides, they do not take into account the positional prior in chest X-rays.

In this work, we propose a \textbf{P}osition-guided \textbf{P}rompt learning method for \textbf{A}nomaly \textbf{D}etection in chest X-rays (\textbf{PPAD}). \our leverages learnable text prompt and image prompt to minimize the gap between pre-training data and task-specific data. The learnable prompts are inserted into the inputs before being fed into the encoders.
Besides, inspired by the fact that experts diagnose chest X-rays by carefully examining distinct lung regions, we leverage the positional prior in chest X-rays to introduce position-guided prompts. Through the position-guided prompts, the model can focus on various regions, simulating the diagnostic process of experts.

Furthermore, we propose a \textbf{S}tructure-preserving \textbf{A}nomaly \textbf{S}ynthesis method (\textbf{SAS}) in the training phase to cope with the issue that learning solely from normal samples may limit model's discrimination capability.
Previous anomaly synthesis methods~\cite{fpi,li2021cutpaste,pii,sato2023anatomy} are primarily focused on cutting from one normal regular region and pasting to another to create synthetic anomaly. However, these methods distort organ structures, potentially leading to model overfitting to this scenario. On the contrary, we utilize Gamma correction on the input image within a random irregular mask to simulate lesions while preserving the structure of lung.
The gamma value in Gamma correction associates with distance to mask's boundary, which allows for seamless anomaly insertion.
Experimental results indicate that SAS generates more authentic anomalies, which is beneficial for anomaly detection in chest X-rays. Besides, the proposed PPAD obtains state-of-the-art performance on three chest X-ray anomaly detection datasets.

\begin{figure}[t]
    \centering
    \includegraphics[width=1.0\textwidth]{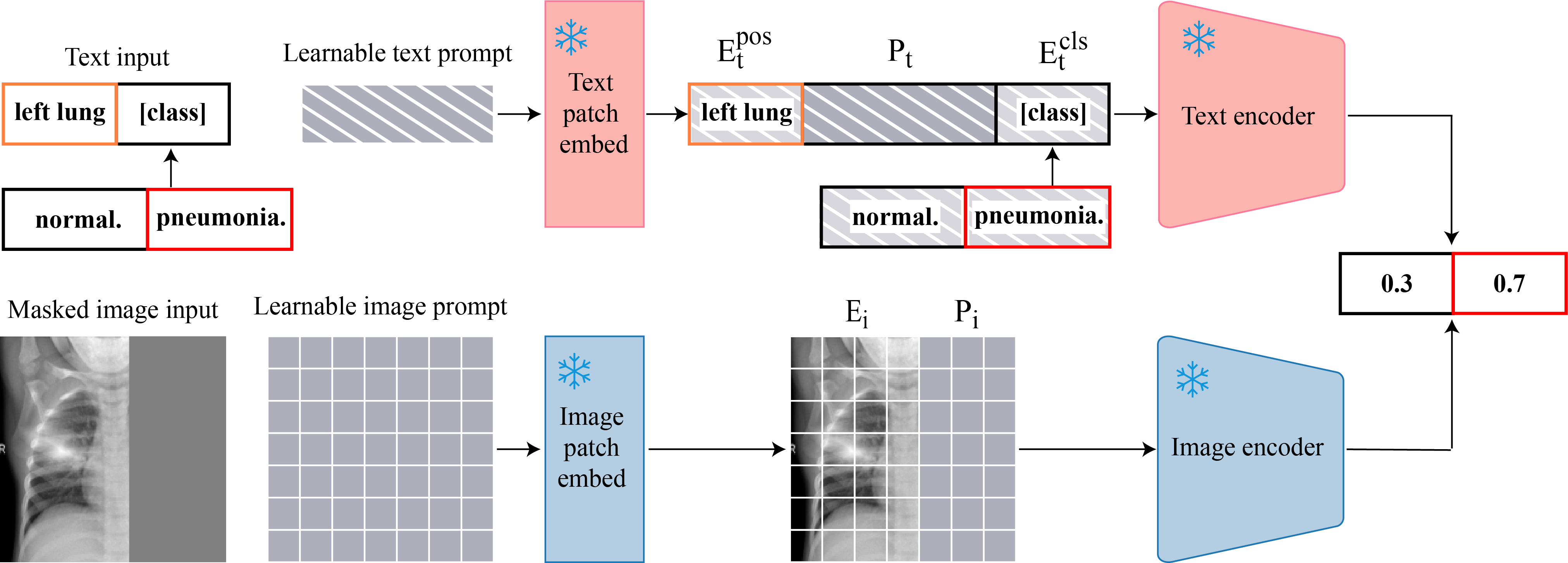}
        \caption{The pipeline of our proposed \our. The main idea of the \our is to adapt text data and image data using learnable prompts. The rest of the model is frozen. Four positional prompts are optional. Taking the ``left lung'' for example, \our incorporates ``left lung'' as the position prompt embeddings $E^{pos}_t$. The learnable text prompt $P_{t}$ is insert between position prompt embeddings $E^{pos}_t$ and class embeddings $E^{cls}_t$. Image input is either the synthetic anomaly or the normal image. The right lung region of the input image embedding $E_i$ is replaced by the learnable image prompt $P_{i}$.}
    \label{fig:main_pipeline}
\end{figure} 

\section{Method}

An overview of the proposed \our is depicted in Fig.~\ref{fig:main_pipeline}. We employ position-guided prompts to adapt both the text and image inputs before feeding them into the encoders. The classification word of text input is substituted with a class name, either ``normal'' or ``pneumonia''. Subsequently, text features and image features are respectively extracted through the text encoder and image encoder. Finally, the prediction probability is computed by applying cosine similarity between the text and image features, followed by a softmax function.

\subsection{Position-guided prompt learning.} 

We leverage the position prior of the \chest to introduce a Position-guided Prompt learning method for \chest Anomaly Detection (PPAD). Specifically, we divide the image into four sections: left lung, right lung, upper lung, and lower lung. Taking the left lung as an example (see Fig.~\ref{fig:main_pipeline}), we initiate the text input with ``left lung''. For the image input, we focus on the left lung within the image, masking the right lung with a binary mask $M$. We mask the right lung to prevent it from influencing the result. 
Subsequently, we introduce learnable text prompt $P_{t}$ and image prompt $P_{i}$. The text input is converted into token embeddings by the text patch embed module. We separate it into position prompt embeddings $E_t^{pos}$ and class embeddings $E_t^{cls}$. The learnable text prompt is inserted between $E_t^{pos}$ and $E_t^{cls}$. The text input $E_{text}$ of the text encoder emerges as a combination, denoted as:
\begin{equation}
   E_{text} = E_t^{pos} \oplus P_{t} \oplus E_t^{cls},
\end{equation}
where $\oplus$ symbolizes embedding concatenation.
The masked image is divided into non-overlapping patches and is further tokenized into embeddings $E_i$, by the image patch embed module. We replace the image patches within the mask $M$ with the learnable image prompt patches. The image input $E_{image}$ of the image encoder is computed as:
\begin{equation}
    E_{image} = E_i \odot M + P_{i} \odot (1-M), 
\end{equation}
where $\odot$ symbolizes element-wise multiplication.
The learnable prompts have the capacity to adapt the training data to the frozen pre-trained CLIP-based model.
Besides, employing positional guidance allows the model to focus on the guided region, simulating the diagnosis results obtained by experts who observe lesions in various regions.

\begin{figure}[t]
    \centering
    \includegraphics[width=1.0\textwidth]{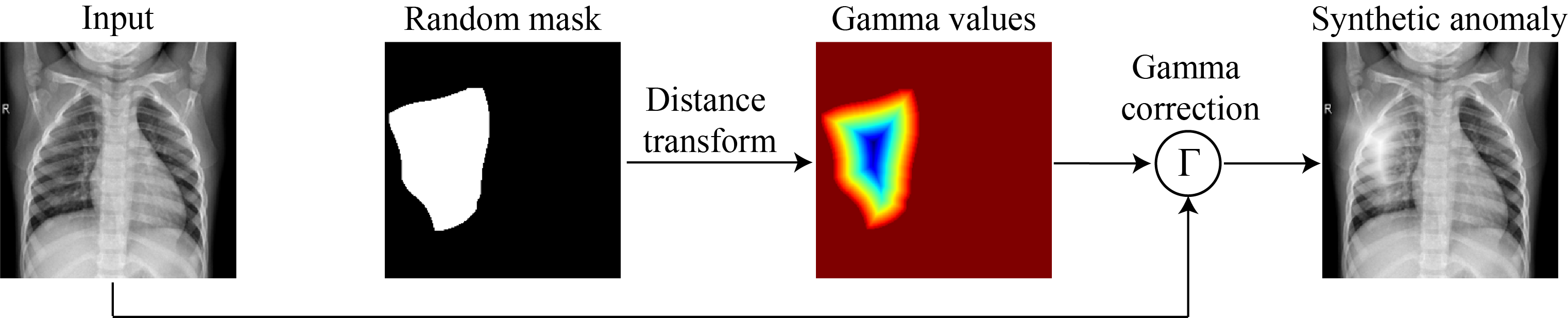}
    \caption{The illustration of the proposed SAS. SAS applies distance transform within a random mask to create smoothed gamma values. Then, synthetic anomaly is generated via Gamma correction applied to the input normal image.}
    \label{fig:pseudo_pipeline}
\end{figure}

\subsection{Structure-preserving anomaly synthesis}

We propose a Structure-preserving Anomaly Synthesis method (SAS) during the training phase 
to enhance the model's discriminative capability. The process is depicted in Fig.~\ref{fig:pseudo_pipeline}.
Firstly, we generate a random, irregular anomaly mask $M_a$ through several steps: 1) Randomly select ten points following the Perlin noise distribution; 2) Determine the smallest enclosing convex polygon for these points; 3) Randomly replace the edges of polygon with second-order Bézier curves; 4) Fill the interior to achieve the random irregular mask.
After generating the anomaly mask, we utilize a distance transform to calculate the distance $D(x)$ from each point $x$ within the anomaly mask to its nearest boundary point.
Then, smoothed gamma values $\gamma(x)$ on each point $x$ within the anomaly mask $M_a$ are given by the randomly assigned weight $w$, using the formula:
\begin{equation}
\label{gamma}
    \gamma(x) = 1 +  \frac{D(x)}{\displaystyle\max_{x' \in M_a}D(x')}  * w ,
\end{equation}
where $w > -1$ is a constraint to ensure $\gamma > 0$. Outside the anomaly mask, $\gamma$ is set to 1.
Finally, the synthetic anomaly results from employing gamma correction $ \Gamma(I) = I^\gamma$ on the input image $I$. Inside the mask, the parameter $\gamma$ is dependent on the distance $D$, ensuring that the closer to the mask's boundary, the less pronounced the grayscale value changes. Outside the mask, grayscale remains unchanged. The distance-related gamma correction allows for seamless anomaly insertion and preserves the lung structure. 
Additionally, we generate irregular masks, rather than rectangle masks used in previous methods, to produce more authentic anomalies.

\subsection{Training and inference}
During the training process, we choose a random position prefix, which can be any one among ``left lung'', ``right lung'', ``upper lung'', ``lower lung'', or ``- -'' (- - symbolizes empty). The empty position prefix indicates a view of the entire lung, where the position name is null and the image is unmasked.
We then apply SAS to generate synthetic anomaly with a 50\% probability. We employ the standard Binary Cross-Entropy (BCE) loss as the loss function. It is important to note that we only optimize the learnable text prompt and image prompt, keeping the other pre-trained parameters frozen.
\par
During the inference phase, we evaluate each image at four separate positions and the entire view, obtaining five distinct probabilities.  If the maximum value among these probabilities exceeds a threshold $\eta$, we employ this maximum value for classification. Otherwise, we calculate the average of these probabilities as the final outcome.

\begin{table}[t]
  \centering
  \caption{Benchmark results on ZhangLab dataset~\cite{zhanglab} and CheXpert dataset~\cite{chexpert}. Top group: results of previous SOTA methods. Bottom group: results of CLIP-based methods and prompt learning methods. The best results in each group are denoted with \textbf{bold}. Methods marked with an asterisk (*) are reproduced in this work. Other results are either from their original paper or from SQUID~\cite{squid}. Mean and standard deviation are the results of five independent experiments.}

  \label{tab:zhang}
  \setlength{\tabcolsep}{1mm}
   \resizebox{1\linewidth}{!}{
  \begin{tabular}{l c | ccc |ccc}
    \toprule

\multirow{2}{*}{Method}  & \makebox[0.05\textwidth]{} & \multicolumn{3}{c|}{ZhangLab dataset~\cite{zhanglab}}  & \multicolumn{3}{c}{CheXpert dataset~\cite{chexpert}}  \\
\cmidrule(l){3-8}
 & & ACC (\%)$\uparrow$ & AUC (\%)$\uparrow$ & F1 (\%)$\uparrow$ & ACC (\%)$\uparrow$ & AUC (\%)$\uparrow$ & F1 (\%)$\uparrow$  \\
\midrule
MemAE~\cite{gong2019memorizing} & & 56.5$\pm$1.1  & 77.8$\pm$1.4 & 82.6$\pm$0.9 & 55.6$\pm$1.4 & 54.3$\pm$4.0 & 53.3$\pm$7.0 \\
SALAD~\cite{salad} & & 75.9$\pm$0.9 & 82.7$\pm$0.8 & 82.1$\pm$0.3  & -  & -  & - \\
CutPaste~\cite{li2021cutpaste} & & 64.0$\pm$6.5 & 73.9$\pm$3.9 & 72.3$\pm$8.9  & 62.7$\pm$2.0 & 65.5$\pm$2.2 & 60.3$\pm$4.6\\
 PANDA~\cite{reiss2021panda} & & 65.4$\pm$1.9 & 65.7$\pm$1.3 & 66.3$\pm$1.2  & 66.4$\pm$2.8 & 68.6$\pm$0.9 & 65.3$\pm$1.5 \\
M-KD~\cite{mkd} & & 69.1$\pm$0.2 & 74.1$\pm$2.6 & 62.3$\pm$8.4 & 66.0$\pm$2.5  & 69.8$\pm$1.6 & 63.6$\pm$5.7 \\
 IF 2D~\cite{if2d} & & 76.4$\pm$0.2 & 81.0$\pm$2.8 & 82.2$\pm$2.7 & -  & -  & - \\
 SQUID~\cite{squid} & & \textbf{80.3$\pm$1.3} & \textbf{87.6$\pm$1.5} & \textbf{84.7$\pm$0.8}  & \textbf{71.9$\pm$3.8} & \textbf{78.1$\pm$5.1} & \textbf{75.9$\pm$5.7} \\
\midrule
 CheXzero*~\cite{chexzero} & & 83.0  & 92.7  & 87.5  & 77.4 & 87.7   & 79.7  \\
 Xplainer*~\cite{xplainer} &  & 78.2  & 89.9  & 85.0  & 75.6 & 83.6   & 75.6  \\
 CoOp*~\cite{zhou2022learning} & & 84.6$\pm$1.7 & 94.6$\pm$1.1 & 88.6$\pm$1.0 & 80.4$\pm$1.8 & 87.0$\pm$1.3 & 79.2$\pm$2.2 \\
 MaPLe*~\cite{khattak2023maple} & & 86.1$\pm$1.0 & 95.1$\pm$1.2 & 89.5$\pm$0.5 & 79.5$\pm$0.8 & 86.3$\pm$1.0 & 79.6$\pm$1.2 \\
\our (Ours) &  & \textbf{89.4$\pm$0.6} & \textbf{96.7$\pm$0.4} & \textbf{91.8$\pm$0.5} & \textbf{82.7$\pm$0.6} & \textbf{88.5$\pm$0.9} & \textbf{82.0$\pm$0.8} \\

    \bottomrule
    
   \end{tabular} 
    }
\end{table}

\section{Experiments}

\subsection{Datasets and implementation details}

\subsubsection{Datasets.} We evaluate our method on three public \chest datasets: 1) ZhangLab Chest X-ray dataset\footnote{\url{https://data.mendeley.com/datasets/rscbjbr9sj/3}}~\cite{zhanglab}, 2) Stanford CheXpert dataset\footnote{\url{https://stanfordmlgroup.github.io/competitions/chexpert}}~\cite{chexpert}, 3) VinBigData Chest X-ray Abnormalities Detection Challenge dataset (VinDr-CXR)\footnote{\url{https://www.kaggle.com/c/vinbigdata-chest-xray-abnormalities-detection}} \cite{vindr}. We employ the dataset setup and evaluation metrics from SQUID~\cite{squid} for the ZhangLab and CheXpert datasets, while those from DDAD~\cite{ddad} are used with the VinDr-CXR dataset. It is important to note that both Xplainer~\cite{xplainer} and our baseline method (CheXzero~\cite{chexzero}), are pre-trained on the MIMIC-CXR dataset~\cite{johnson2019mimic}, which is unrelated to the datasets utilized in our task. Details of datasets setup are described in the Supplementary Material. 

\begin{table}[t]
  \centering
  
  \caption{Benchmark results on VinDr-CXR dataset~\cite{vindr}. The results of top group are either from their respective original paper or from DDAD~\cite{ddad}.}
  
  \label{tab:rsna}
  \setlength{\tabcolsep}{1mm}
   \resizebox{1.0\linewidth}{!}{
  \begin{tabular}{l c| cc cc cc cc c}

    \toprule
Method           & \makebox[0.01\textwidth]{}  & CutPaste~\cite{li2021cutpaste} & \makebox[0.005\textwidth]{} & IGD~\cite{chen2022deep} & \makebox[0.005\textwidth]{} & AMAE~\cite{amae} & \makebox[0.005\textwidth]{} & Baugh \etal~\cite{baugh2023many} & \makebox[0.005\textwidth]{} & DDAD~\cite{ddad} \\
\midrule
AUC (\%)$\uparrow$ & & 70.2 && 59.2 && 74.2  && 76.6 && \textbf{78.2}    \\

AP (\%)$\uparrow$ && 69.8 && 58.7 && 72.9 && \textbf{78.4} && 74.6    \\

\midrule
Method            & & CheXzero*~\cite{chexzero} && Xplainer*~\cite{xplainer} && CoOp*~\cite{zhou2022learning} && MaPLe*~\cite{khattak2023maple} && \our (Ours) \\
\midrule
AUC (\%)$\uparrow$ && 77.6   && 76.5  && 79.9$\pm$2.1 && 74.2$\pm$1.3 && \textbf{81.9$\pm$0.9}     \\

AP (\%)$\uparrow$ &&  78.4 && 78.2 && 80.9$\pm$2.1 && 75.7$\pm$1.1 & & \textbf{82.1$\pm$1.1}   \\
    
    \bottomrule

   \end{tabular} 
    }
\end{table}

\noindent
\subsubsection{Implementation details.} 
We implement the proposed \our, together with CoOp~\cite{zhou2022learning} and MaPLe~\cite{khattak2023maple}, based on CheXzero~\cite{chexzero} with the proposed SAS employed as the anomaly synthesis method.  The total training epoch is set to 100. By default, the hyperparameter $w$ involved in Eq.~\ref{gamma} is randomly chosen from \{-0.999, -0.99, 2, 3\} to introduce diverse synthetic anomalies. The threshold $\eta$ is set to 0.8. The few-shot evaluation protocol established in CLIP~\cite{clip} is adopted, which uses 64 shots randomly selected from normal training sets during training. We report the mean and standard deviation of the proposed \our and the reproduced methods, computed over five runs.

\subsection{Main results}
Table~\ref{tab:zhang} and table~\ref{tab:rsna} are divided into two groups for comparison.
The first group compares \our with some SOTA methods~\cite{gong2019memorizing,salad,li2021cutpaste,reiss2021panda,mkd,if2d,squid,chen2022deep,amae,baugh2023many,ddad}. The second group, \our compares with both CLIP-based methods~\cite{chexzero,xplainer} and prompt learning methods~\cite{zhou2022learning,khattak2023maple}.
On the ZhangLab dataset~\cite{zhanglab}, \our outperforms previous SOTA method SQUID~\cite{squid} by 9.1\%, 9.1\%, 7.1\% in ACC, AUC, and F1, respectively. When compared to the CLIP-based methods CheXzero~\cite{chexzero} and Xplainer~\cite{xplainer}, \our achieves 6.4\%, 4.0\%, 4.3\% and 11.2\%, 6.8\%, 6.8\% in ACC, AUC, and F1 improvement. Furthermore, in comparison to the prompt learning methods CoOp~\cite{zhou2022learning} and MaPLe~\cite{khattak2023maple}, \our achieves improvements of 4.8\%, 2.1\%, 3.2\% and of 3.3\%, 1.6\%, 2.3\% in ACC, AUC, and F1, respectively. The same conclusion can be drawn on the other two datasets. On the CheXpert dataset~\cite{chexpert}, we surpass SQUID by 10.8\%, 10.4\%, and 6.1\% in ACC, AUC, F1, respectively, and outperform CheXzero by 5.3\%, 0.8\%, and 2.3\%. On the VinDr-CXR dataset~\cite{vindr}, \our outperforms DDAD~\cite{ddad} and CheXzero by 3.7\%, 7.5\% and 4.3\%, 3.7\% in AUC, AP, respectively.

\begin{figure}[t]
\centering
\subfloat[{\scriptsize Input}]{
\begin{minipage}[b]{0.16\linewidth}
\includegraphics[width=1.0\linewidth]{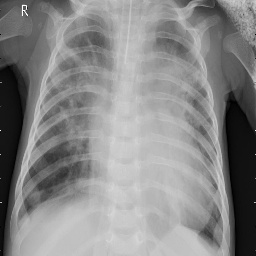}\vspace{1pt}
\includegraphics[width=1.0\linewidth]{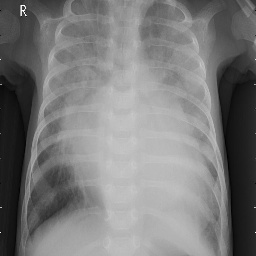}\vspace{1pt}
\end{minipage} \hspace{-8pt}
} 
\subfloat[{\scriptsize Entire lung}]{
\begin{minipage}[b]{0.16\linewidth}
\includegraphics[width=1\linewidth]{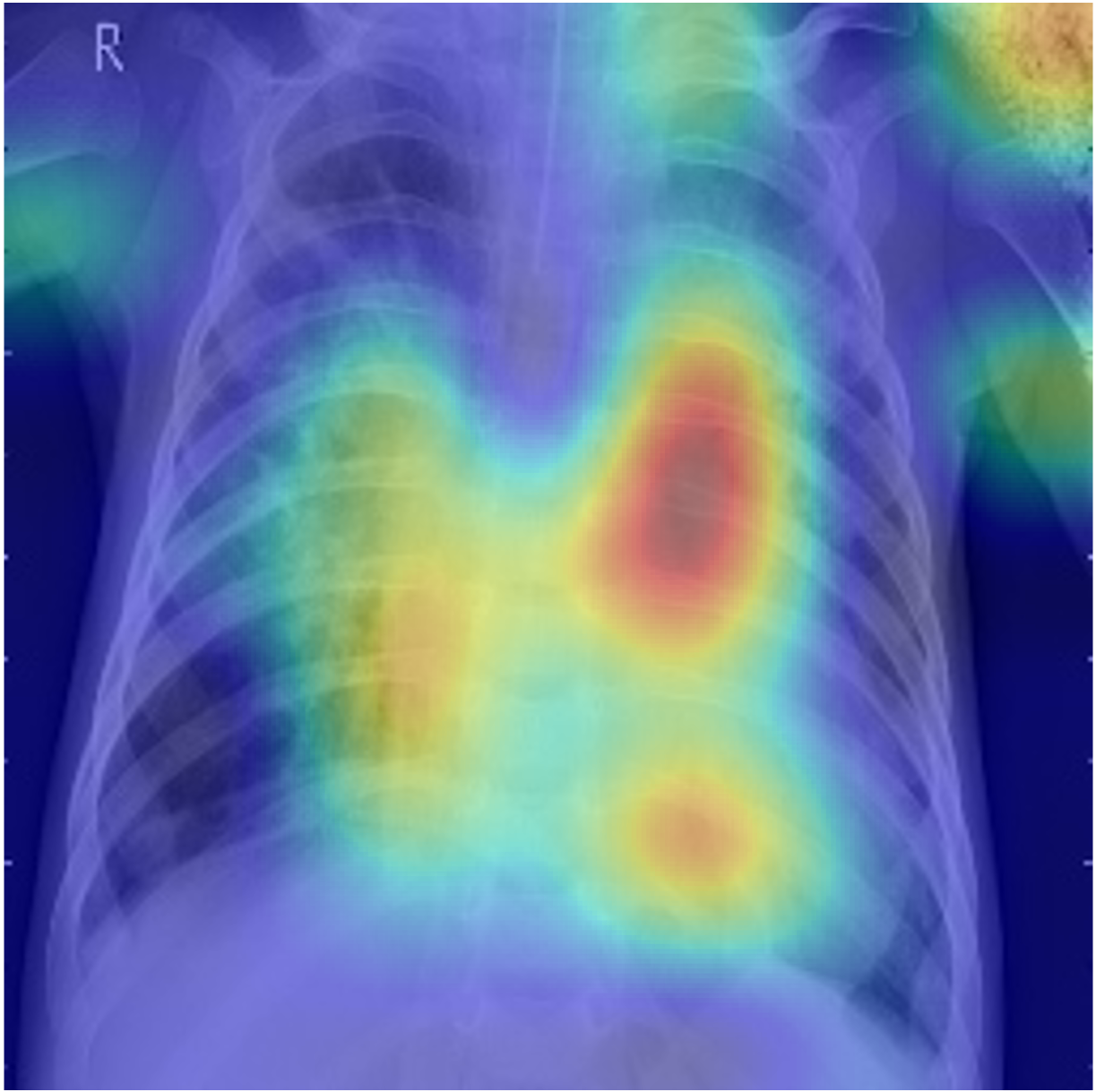}\vspace{1pt}
\includegraphics[width=1\linewidth]{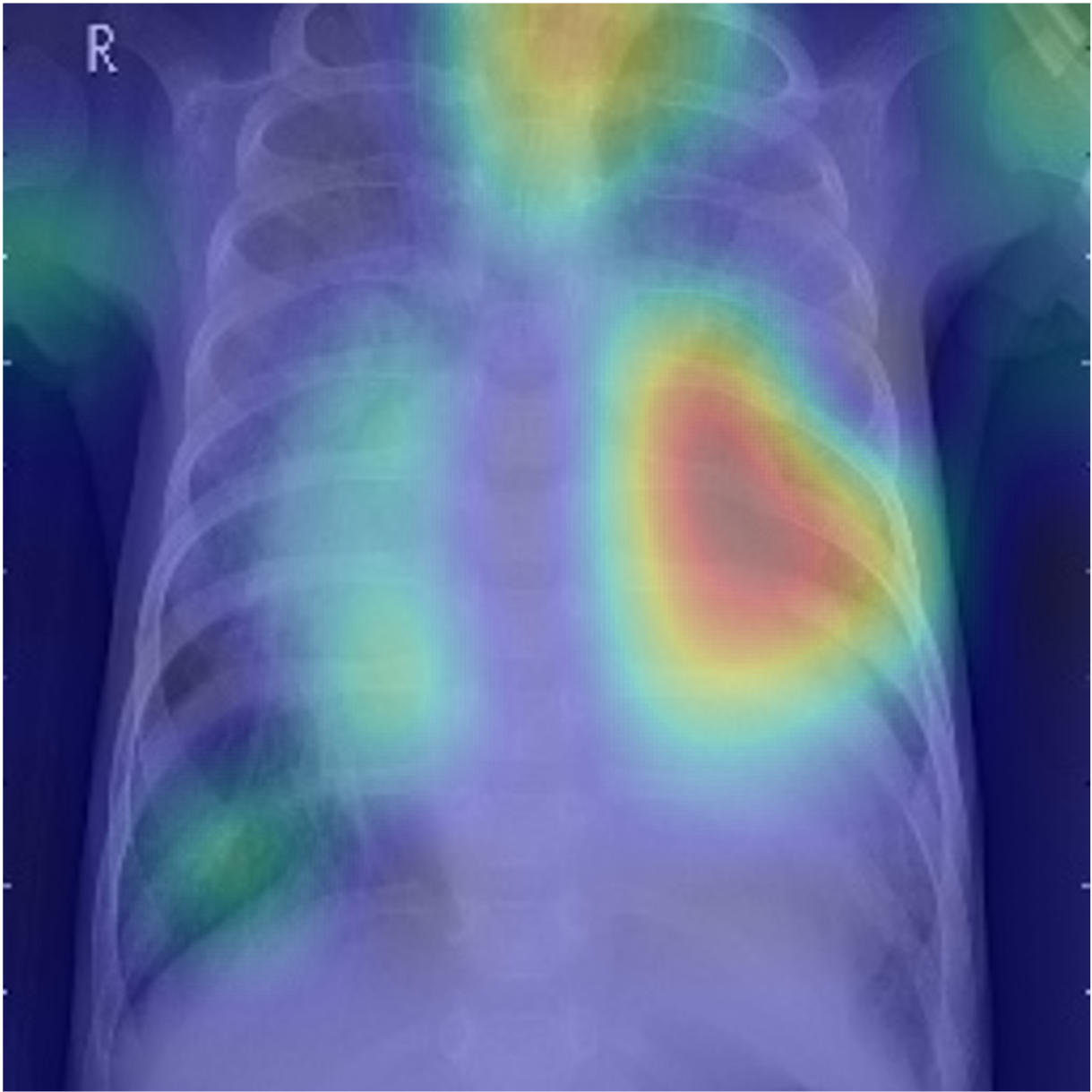}\vspace{1pt}
\end{minipage} \hspace{-8pt}
}
\begin{minipage}{0.01\linewidth}
  \centering
  \tikz[remember picture,overlay] \draw[line width=0.3mm, dashed] (0.08,.0\textheight) -- (0.08,.2\textheight);
\end{minipage}%
\subfloat[{\scriptsize Left lung}]{
\begin{minipage}[b]{0.16\linewidth}
\includegraphics[width=1\linewidth]{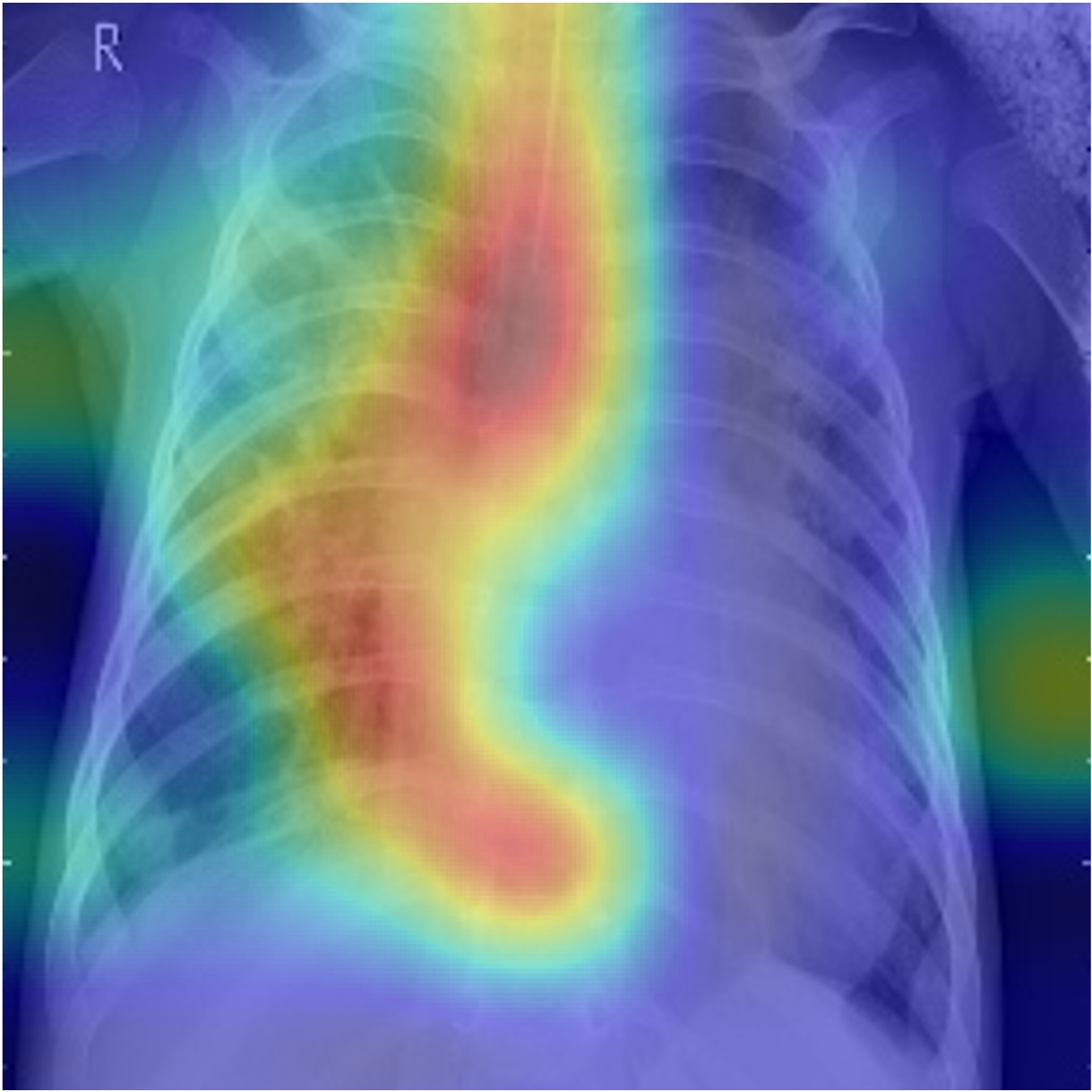}\vspace{1pt}
\includegraphics[width=1\linewidth]{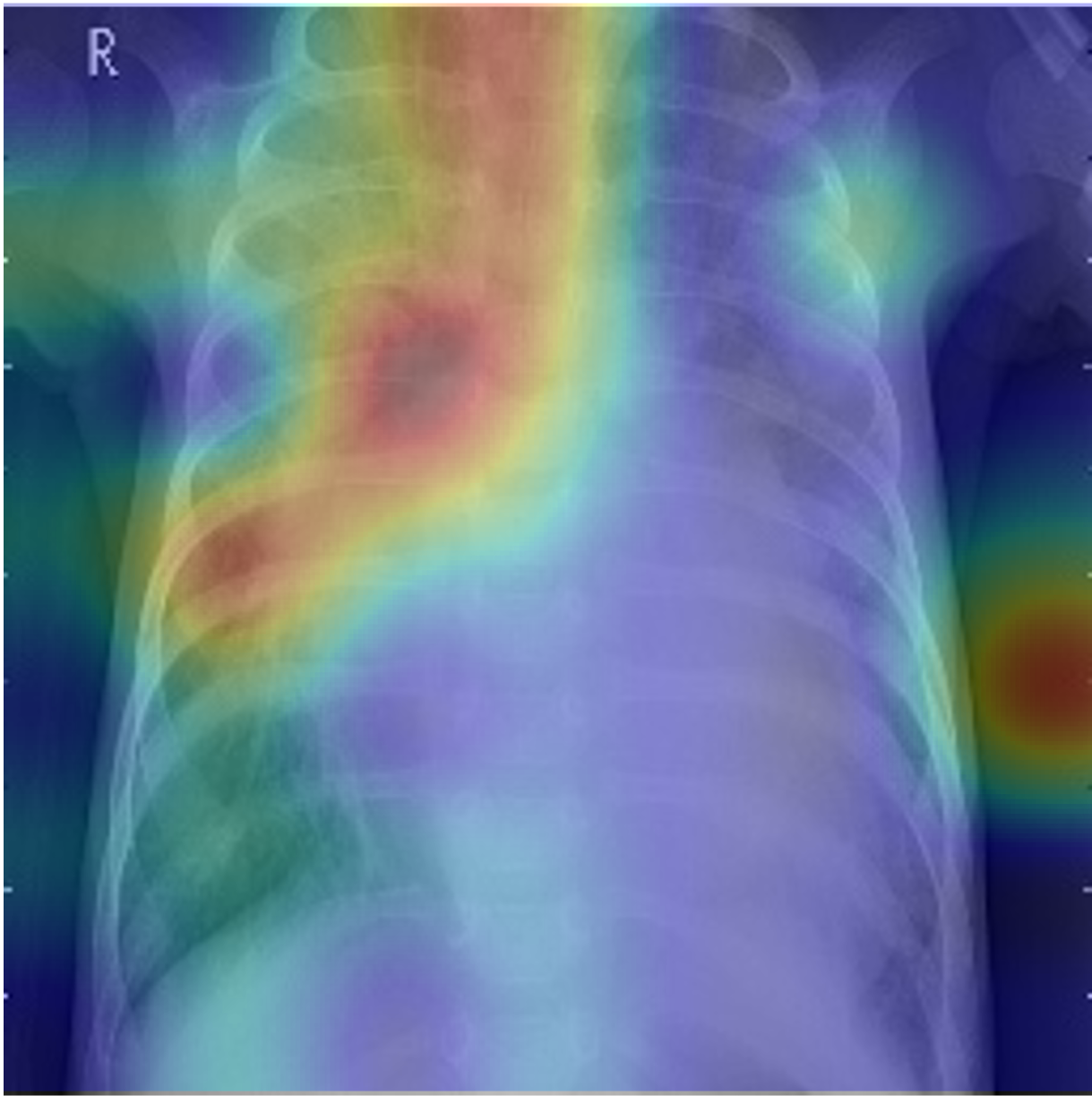}\vspace{1pt}
\end{minipage} \hspace{-8pt}
}
\subfloat[{\scriptsize Right lung}]{
\begin{minipage}[b]{0.16\linewidth}
\includegraphics[width=1\linewidth]{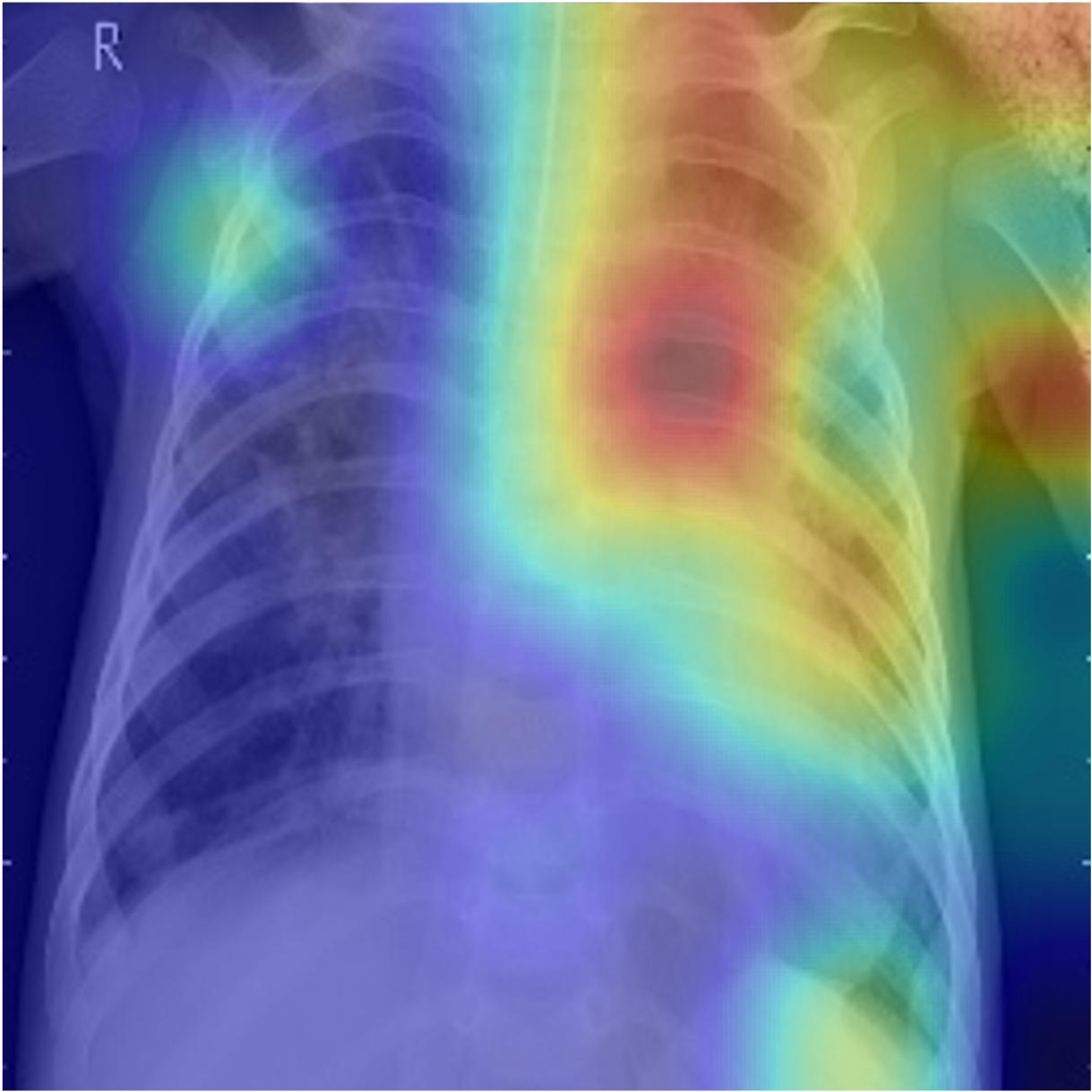}\vspace{1pt}
\includegraphics[width=1\linewidth]{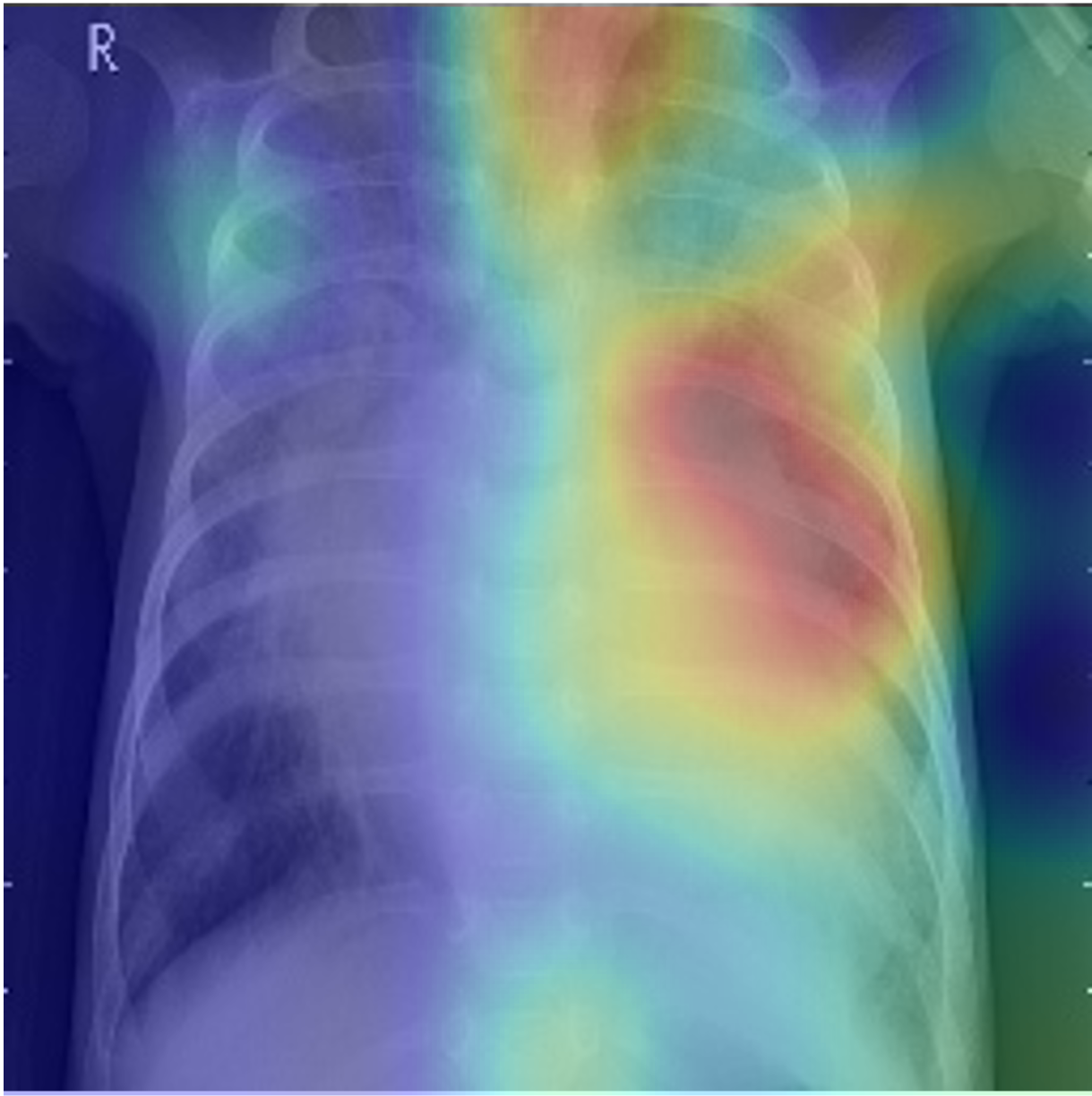}\vspace{1pt}
\end{minipage} \hspace{-8pt}
}
\subfloat[{\scriptsize Upper lung}]{
\begin{minipage}[b]{0.16\linewidth}
\includegraphics[width=1\linewidth]{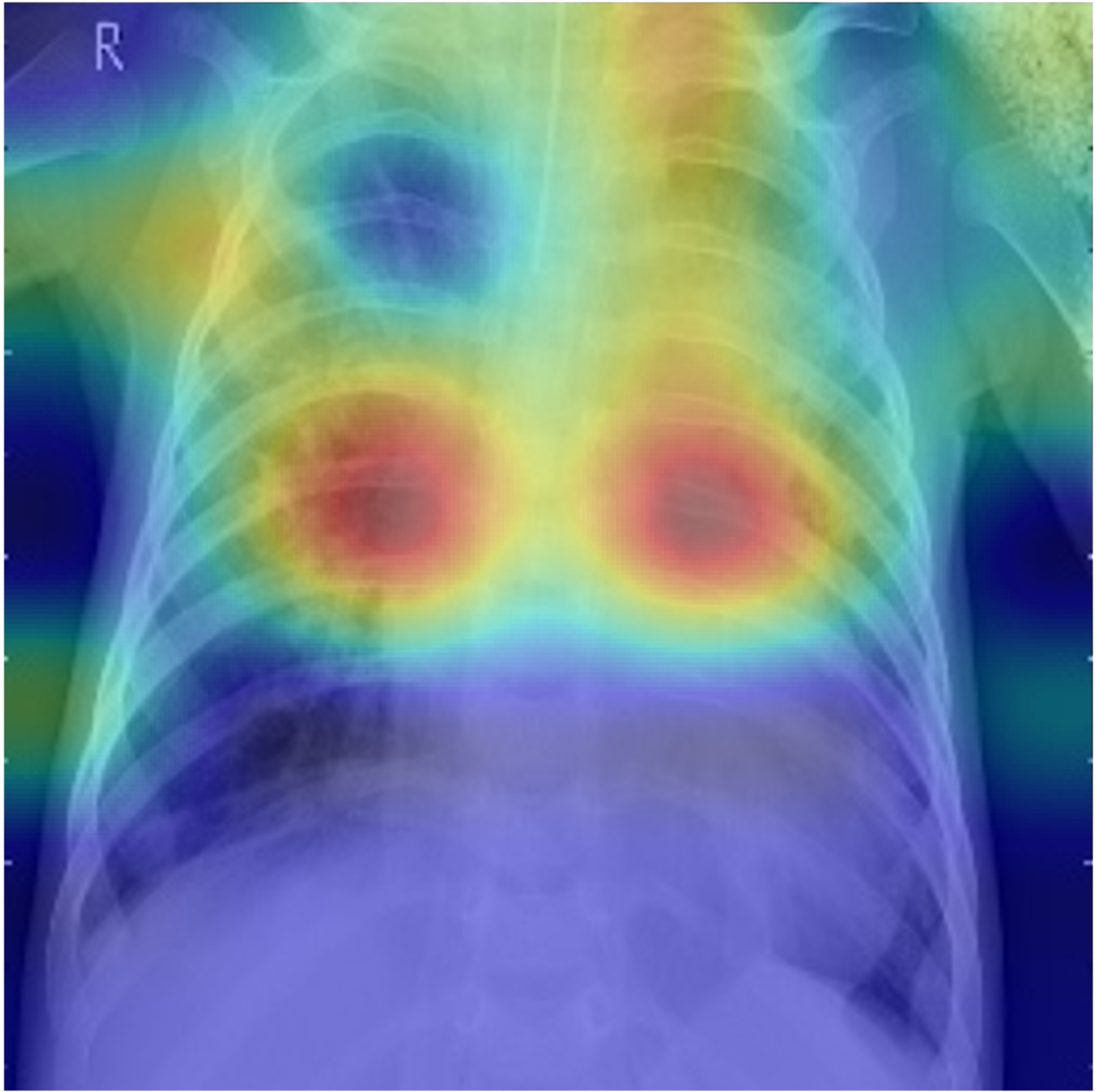}\vspace{1pt}
\includegraphics[width=1\linewidth]{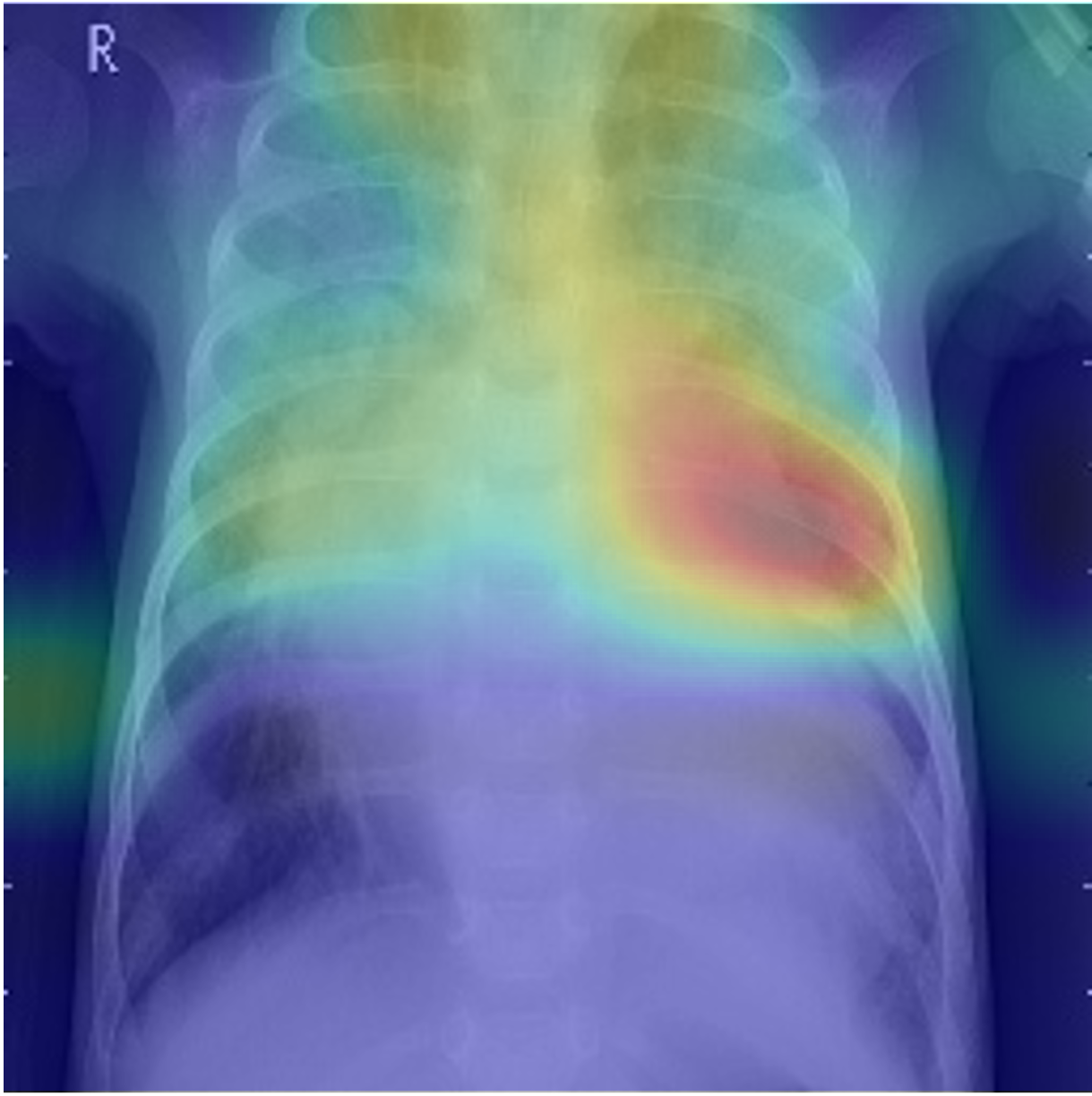}\vspace{1pt}
\end{minipage} \hspace{-8pt}
}
\subfloat[{\scriptsize Lower lung}]{
\begin{minipage}[b]{0.16\linewidth}
\includegraphics[width=1\linewidth]{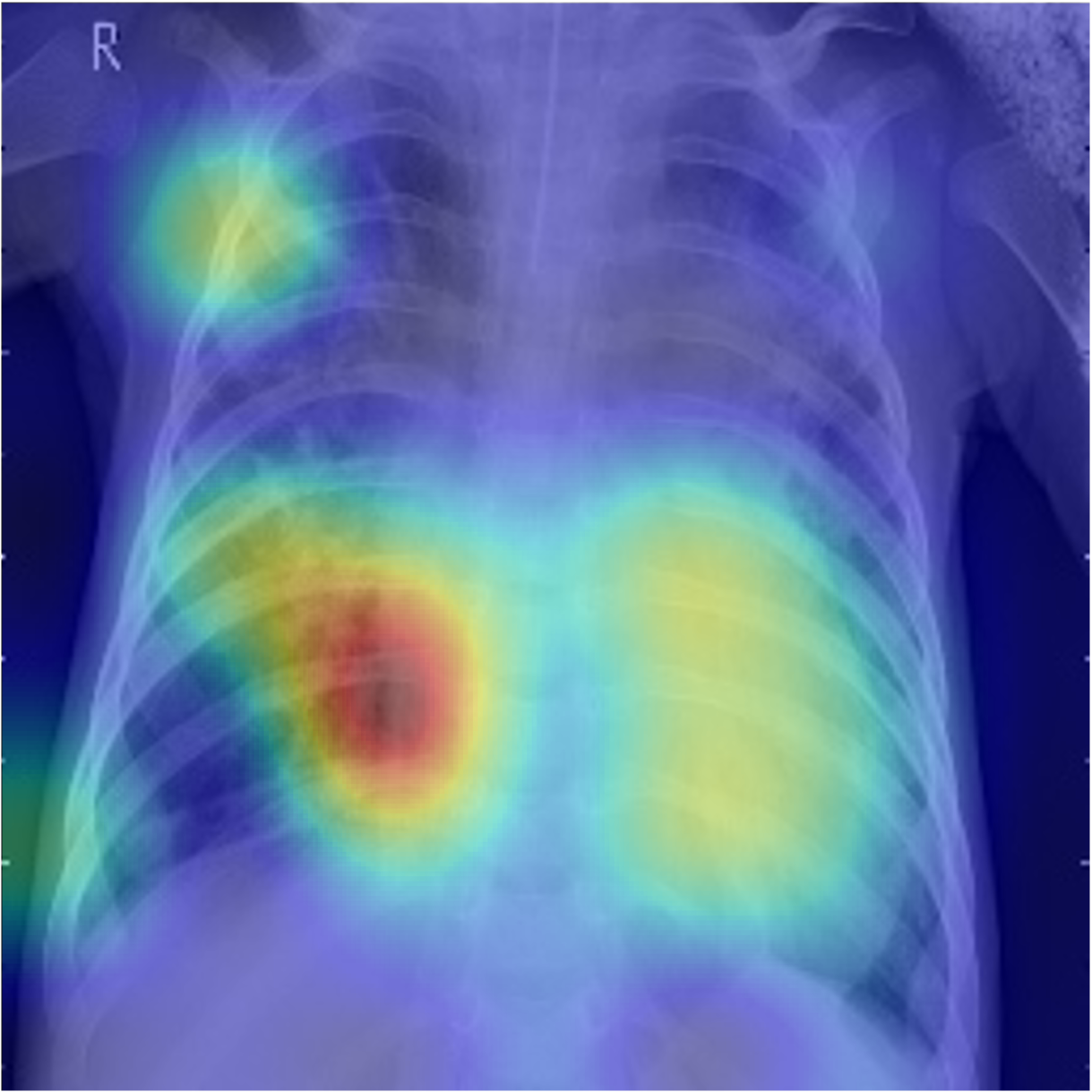}\vspace{1pt}
\includegraphics[width=1\linewidth]{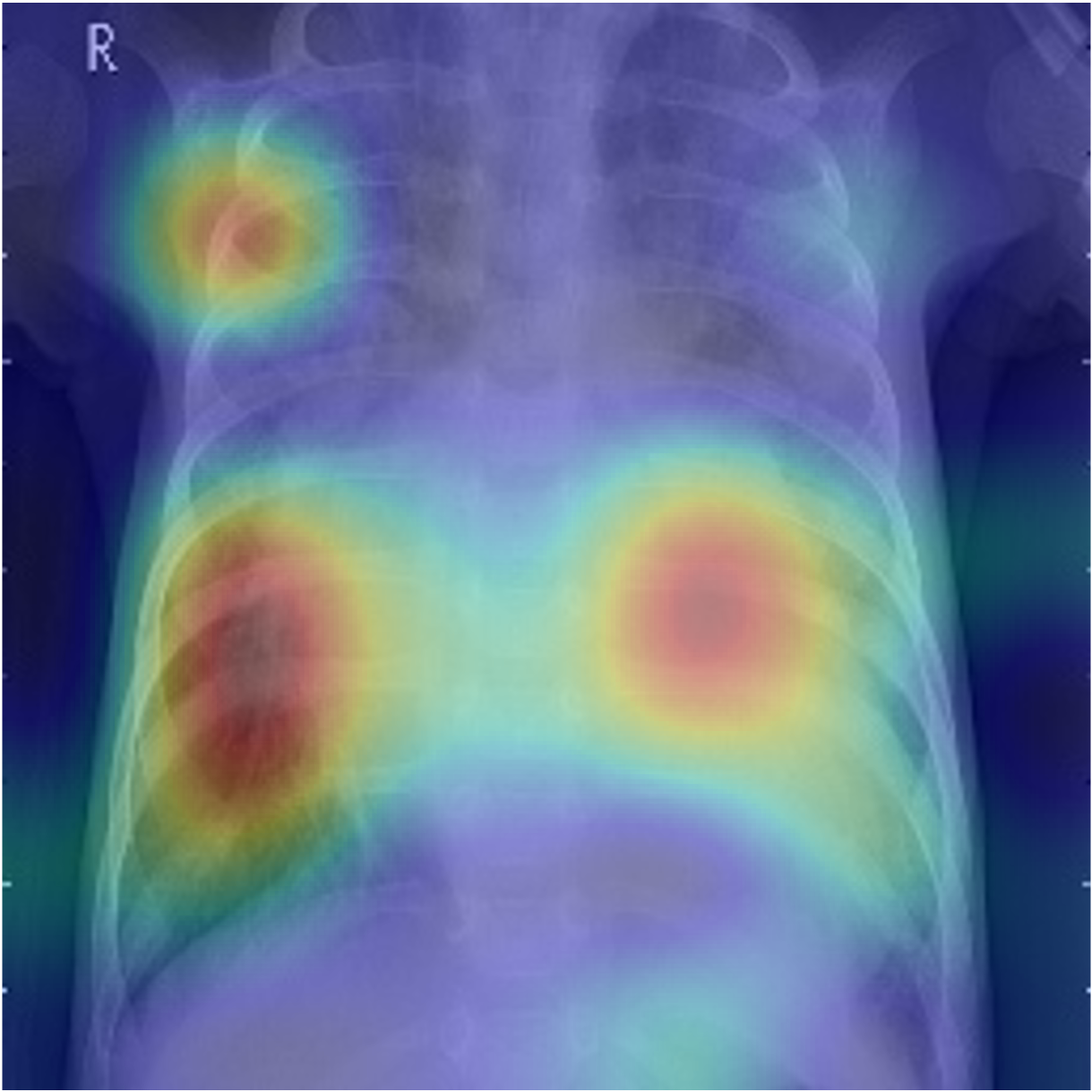}\vspace{1pt}
\end{minipage} \hspace{-8pt}
}
\caption{The visualization of CAM guided by the entire view (the second column) and various position prompts (last four columns). }

\label{fig:vis_position}
\end{figure}

\begin{table}[t]
\caption{Ablation study on Zhanglab dataset~\cite{zhanglab} for the contribution of each position-guided prompt in \our.}

\label{tab:ablation}
  \centering
  \begin{adjustbox}{max width=\textwidth}

  \begin{tabular}{l c |c c c c c c c}
    \toprule
   Method  & \makebox[0.02\textwidth]{} &   {\small ACC (\%)$\uparrow$} & \makebox[0.01\textwidth]{} & {\small AUC (\%)$\uparrow$}  & \makebox[0.01\textwidth]{} & {\small F1 (\%)$\uparrow$}  & \makebox[0.01\textwidth]{} & {\small AP (\%)$\uparrow$ } \\ 
    \midrule 
    Baseline (CheXzero~\cite{chexzero})  &  & 83.0  & &  92.7 & & 87.5 & & 95.4  \\
    + Learnable text prompt (CoOp~\cite{zhou2022learning})  & & 84.6$\pm$1.7  & &  94.6$\pm$1.1 & & 88.6$\pm$1.0 & & 96.7$\pm$0.9 \\
    + Position-guided text prompt  &  & 86.7$\pm$2.2 & &   95.9$\pm$0.5 & & 89.3$\pm$2.5 & &97.7$\pm$0.3 \\
    + Position-guided text and image prompt  &   & \textbf{89.4$\pm$0.6} & & \textbf{96.7$\pm$0.4} & & \textbf{91.8$\pm$0.5} & &\textbf{98.0$\pm$0.3} \\
  \bottomrule

\end{tabular}
\end{adjustbox}

\end{table}

\subsection{Ablation studies}

\subsubsection{Visualization of position-guided impact.} The impact of position guidance is visualized through the Class Activation Mapping (CAM) in Fig.~\ref{fig:vis_position}. 
In the first row, the entire lung view fails to detect anomalies in the left region. However, specific views focused on the left, upper, and lower sections of the lung successfully identify these anomalies. In the second row, the left lung view detects anomalies that are missed in the entire lung view. These outcomes indicate the importance of utilizing diverse position prompts.
\par
\noindent
\subsubsection{Quantitative analysis of position-guided prompts.} 
As presented in Table~\ref{tab:ablation}, the baseline method CheXzero~\cite{chexzero} achieves an ACC of 83.0\%. The introduction of a learnable text prompt (CoOp~\cite{zhou2022learning}) improves the ACC by 1.6\%. Introducing the position-guided text prompt further improves the ACC by 2.1\%. Subsequently, the integration of the position-guided learnable image prompt leads to an extra increase of 2.7\%. These results demonstrate the contribution of each position-guided prompt in \our.
\par
\noindent
\subsubsection{Quantitative analysis of the proposed SAS.} 
As detailed in Table~\ref{tab:compare_pseudo}, the proposed SAS outperforms other anomaly synthesis methods including FPI~\cite{fpi}, PII~\cite{pii}, CutPaste~\cite{li2021cutpaste}, and AnatPaste~\cite{sato2023anatomy}, all of which are based on the \our framework.
The SAS synthesises anomalies that resemble natural diseases, thereby improving the detection of anomalies in chest X-rays. The visualization results of different synthetic anomalies are shown in Supplementary Material.
\par
\noindent
\subsubsection{Ablation with varied numbers of training shots.} 
We perform an ablation study focusing on the number of shots in the training dataset in the Supplementary Material. To balance the effectiveness and training time, we choose 64 shots during the training phase.

\begin{table}[t]
  \centering
  \caption{
  Ablation study on applying various anomaly synthesis methods to the \our.}
  \label{tab:compare_pseudo}
  \setlength{\tabcolsep}{1mm}
   \resizebox{1\linewidth}{!}{
  \begin{tabular}{l c c c c c c c c}
    \toprule
    \textit{ZhangLab}~\cite{zhanglab}  & \makebox[0.05\textwidth]{}  & ACC (\%)$\uparrow$  & \makebox[0.05\textwidth]{} & AUC (\%)$\uparrow$  & \makebox[0.05\textwidth]{} & F1(\%)$\uparrow$  & \makebox[0.05\textwidth]{}  &  AP(\%)$\uparrow$ \\
    \midrule
    \our + FPI~\cite{fpi} & & 87.0$\pm$0.8 & & 94.2$\pm$0.7 & & 90.0$\pm$0.8 & & 96.2$\pm$0.7 \\
    \our + PII~\cite{pii} & & 88.1$\pm$1.1 & & 95.7$\pm$0.7 & & 90.5$\pm$1.2 & & 97.5$\pm$0.4 \\
    \our + CutPaste~\cite{li2021cutpaste} & & 87.0$\pm$0.8 & & 94.4$\pm$0.8 & & 89.7$\pm$0.6 & & 96.4$\pm$0.6 \\
    \our + AnatPaste~\cite{sato2023anatomy} & & 87.3$\pm$1.1 & & 95.4$\pm$1.3 & & 90.0$\pm$0.7 & & 97.3$\pm$0.8 \\
    
    \our + SAS (Ours)  & & \textbf{89.4$\pm$0.6} & & \textbf{96.7$\pm$0.4} & & \textbf{91.8$\pm$0.5} & & \textbf{98.0$\pm$0.3} \\

    \bottomrule

   \end{tabular} 
    }
\end{table}

\par

\subsection{Discussion}
\subsubsection{``Pneumonia'' enables anomaly detection work.} The reason that the class name ``Pneumonia'' enables anomaly detection work is based on two aspects. First, pneumonia is a prevalent and representative anomaly category in chest X-rays. Second, the inherent capacity of analogical generalization in the CLIP model ensures that the similarity between normal and pathological cases is less than the similarity between pneumonia and pathological cases. This aids in performing anomaly detection by measuring these similarity values. Besides, this analogical generalization capacity may extend the CLIP model to detect rare pathological categories.

\par
\noindent
\subsubsection{Training efficiency.} 
In terms of training parameters, the \our only has 0.046M learnable parameters, which is about 0.03\% of the training parameters in the ViT-B/32 CLIP (151.32M). 
In terms of overall training time, the \our takes about 13 minutes in the 64-shot and 100-epoch setting.
Overall, the proposed \our is efficient for practical deployment.

\par
\noindent
\subsubsection{Limitation.}
Our anomaly synthesis method SAS appears to fall short in simulating specific diseases, such as cardiomegaly and fractures. This shortcoming becomes evident when considering its performance on the CheXpert~\cite{chexpert} dataset, which comprises 12 distinct diseases. The proposed method only achieves a modest improvement of 2.8\% (average across three metrics) over the baseline method CheXzero~\cite{chexzero} when tested on this dataset.

\section{Conclusion}
We propose PPAD, a position-guided prompt learning method for anomaly detection in chest X-rays. PPAD is based on CLIP-based methods to adapt the task-specific data to the pre-training data. Additionally, we introduce SAS for synthetic anomaly. Experiments on three datasets demonstrate that the PPAD achieves state-of-the-art performance. 
In the future, we intend to extend our PPAD method to achieve anomaly localization. Besides, we aim to refine our anomaly synthesis method SAS to incorporate a wider variety of anomalies.

\section*{Supplementary Materials}\label{sec:supplementary}
\begin{table}[ht]
  \centering
  \caption{An overview of the datasets, including the training set, testing set, and the number of anomaly categories.}
  \label{tab:dataset}
   \setlength{\tabcolsep}{1mm}
   \resizebox{1\linewidth}{!}{
  \begin{tabular}{l |c | c |c }

    \toprule
    Dataset   & Training set & Testing set & Anomaly category  \\
    \midrule

    ZhangLab  & 1249 normal images & 234 normal + 390 abnormal images & 2 \\
    CheXpert  & 4499 normal images & 250 normal + 250 abnormal images & 12  \\
    VinDr-CXR  & 4000 normal images & 1000 normal + 1000 abnormal images & 27 \\

    \bottomrule

   \end{tabular} 
    }
\end{table}

\begin{table}[ht]
  \centering
  \caption{
  Additional experimental results of CoOp and MaPLe applying FPI as anomaly synthesis method, with highlighted rows in gray.
  }
  \label{tab:coop}
  \setlength{\tabcolsep}{1mm}
   \resizebox{1\linewidth}{!}{

    \begin{tabular}{l cccc ccccc}

    \toprule
    \textit{ZhangLab} & \makebox[0.08\textwidth]{} & ACC (\%)$\uparrow$ & \makebox[0.08\textwidth]{} & AUC (\%)$\uparrow$  & \makebox[0.08\textwidth]{} & F1(\%)$\uparrow$ & \makebox[0.08\textwidth]{} &  AP(\%)$\uparrow$ \\
    \midrule

    CheXzero & \makebox[0.08\textwidth]{} & 83.0 & \makebox[0.08\textwidth]{} & 92.7 & \makebox[0.08\textwidth]{} & 87.5  & \makebox[0.08\textwidth]{} & 95.4 \\

    \midrule
    CoOp + FPI   & \makebox[0.08\textwidth]{} & 83.6$\pm$2.8 & \makebox[0.08\textwidth]{} & 92.9$\pm$1.0 & \makebox[0.08\textwidth]{} & 87.9$\pm$1.7  & \makebox[0.08\textwidth]{} & 95.6$\pm$0.6 \\
    CoOp + SAS  & \makebox[0.08\textwidth]{} & 84.6$\pm$1.7 & \makebox[0.08\textwidth]{} & 94.6$\pm$1.1 & \makebox[0.08\textwidth]{} & 88.6$\pm$1.0  & \makebox[0.08\textwidth]{} & 96.7 $\pm$0.9 \\

    \midrule
    MaPLe + FPI    & \makebox[0.08\textwidth]{} & 84.4$\pm$1.8 & \makebox[0.08\textwidth]{}  & 93.5$\pm$1.4 & \makebox[0.08\textwidth]{} & 88.3$\pm$1.2 & \makebox[0.08\textwidth]{}   & 96.0$\pm$1.0 \\
    MaPLe + SAS   &\makebox[0.08\textwidth]{}  & 86.1$\pm$1.0 & \makebox[0.08\textwidth]{} & 95.1$\pm$1.2 & \makebox[0.08\textwidth]{} & 89.5$\pm$0.5  & \makebox[0.08\textwidth]{} & 97.1$\pm$0.8 \\
    \midrule
    PPAD + FPI  & \makebox[0.08\textwidth]{} & 87.0$\pm$0.8 & \makebox[0.08\textwidth]{} & 94.2$\pm$0.7 & \makebox[0.08\textwidth]{} & 90.0$\pm$0.8 & \makebox[0.08\textwidth]{}  & 96.2$\pm$0.7 \\
    PPAD + SAS  & \makebox[0.08\textwidth]{} & \textbf{89.4$\pm$0.6}  & \makebox[0.08\textwidth]{} & \textbf{96.7$\pm$0.4} & \makebox[0.08\textwidth]{} & \textbf{91.8$\pm$0.5} &\makebox[0.08\textwidth]{} & \textbf{98.0$\pm$0.3} \\

    \bottomrule

   \end{tabular} 
    }
\end{table}

\begin{table}[ht]
  \centering
  \caption{Ablation study of the proposed PPAD with various numbers of training shots on the ZhangLab dataset. ACC serves as the evaluation metric.}
  \label{tab:shot_num}
  \setlength{\tabcolsep}{1mm}
   \resizebox{1\linewidth}{!}{
  \begin{tabular}{l c  c c c  c c }

    \toprule
    Dataset  & \makebox[0.18\textwidth]{} & 16 shots  & \makebox[0.18\textwidth]{} & 32 shots  & \makebox[0.18\textwidth]{}  & 64 shots  \\
    \midrule
    \textit{ZhangLab}  & \makebox[0.10\textwidth]{} &  87.0$\pm$1.8  & \makebox[0.10\textwidth]{} & 88.6$\pm$0.8  & \makebox[0.10\textwidth]{} & \textbf{89.4$\pm$0.6}  \\

    \bottomrule

   \end{tabular} 
    }
\end{table}

\begin{figure}[htp]
\centering
   
\subfloat[Input]{
\begin{minipage}[b]{0.16\linewidth}
\includegraphics[width=1.0\linewidth]{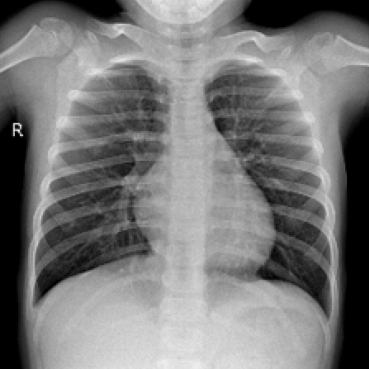}\vspace{0pt}
\includegraphics[width=1.0\linewidth]{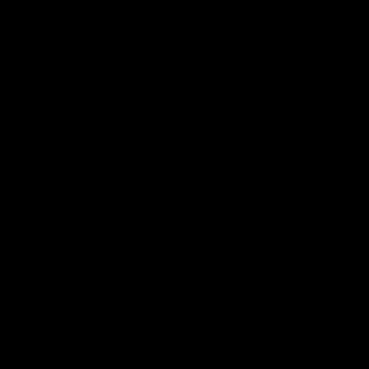}\vspace{0pt}
\includegraphics[width=1.0\linewidth]{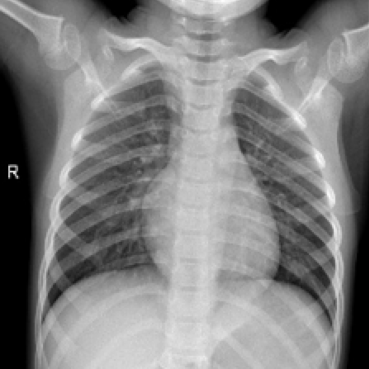}\vspace{0pt}
\includegraphics[width=1.0\linewidth]{org_img_mask.png}\vspace{0pt}
\includegraphics[width=1.0\linewidth]{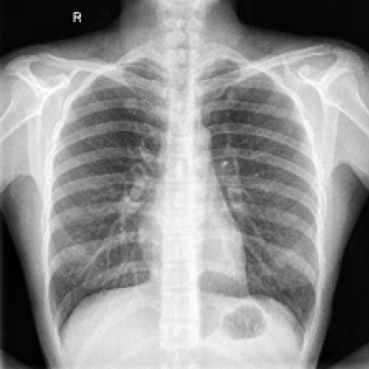}\vspace{0pt}
\includegraphics[width=1.0\linewidth]{org_img_mask.png}\vspace{0pt}
\end{minipage} \hspace{-8pt}
} 
\subfloat[FPI]{
\begin{minipage}[b]{0.16\linewidth}
\includegraphics[width=1.0\linewidth]{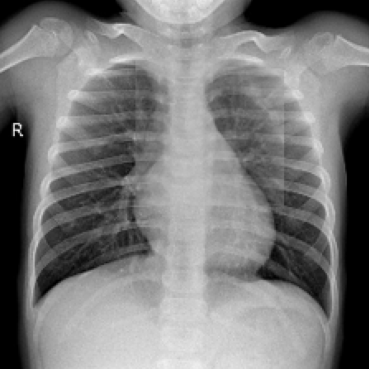}\vspace{0pt}
\includegraphics[width=1.0\linewidth]{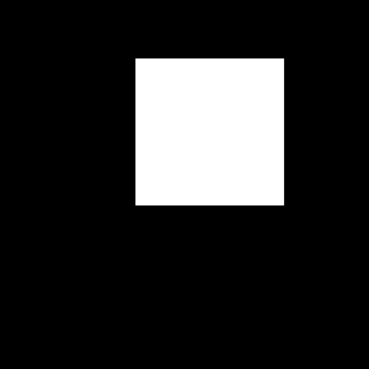}\vspace{0pt}
\includegraphics[width=1.0\linewidth]{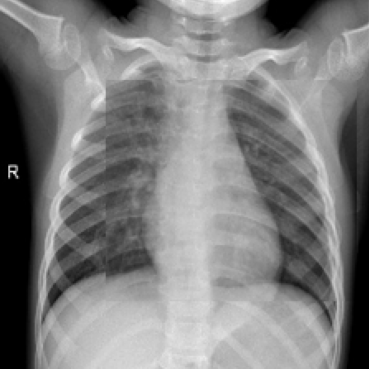}\vspace{0pt}
\includegraphics[width=1.0\linewidth]{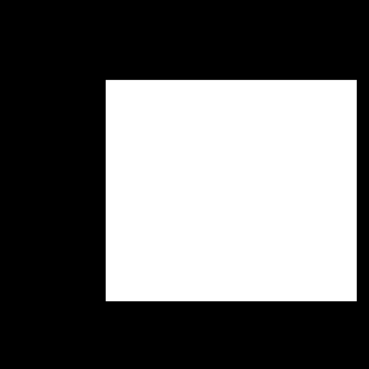}\vspace{0pt}
\includegraphics[width=1.0\linewidth]{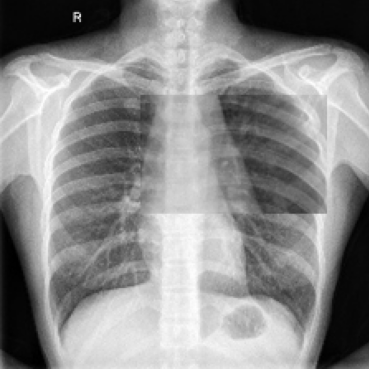}\vspace{0pt}
\includegraphics[width=1.0\linewidth]{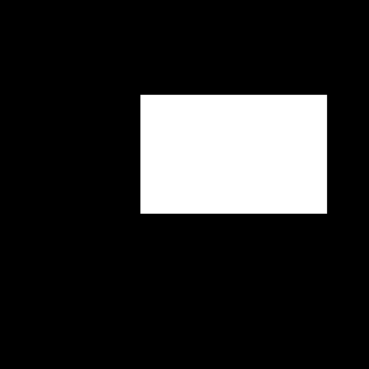}\vspace{0pt}
\end{minipage} \hspace{-8pt}
}
\subfloat[PII]{
\begin{minipage}[b]{0.16\linewidth}
\includegraphics[width=1.0\linewidth]{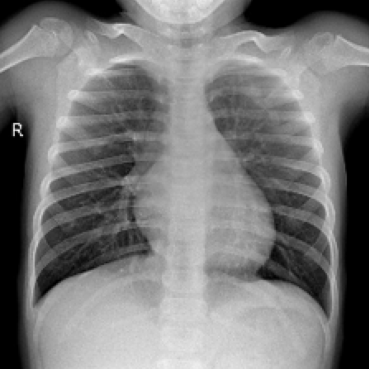}\vspace{0pt}
\includegraphics[width=1.0\linewidth]{NORMAL-28501-0001_fpi_mask.png}\vspace{0pt}
\includegraphics[width=1.0\linewidth]{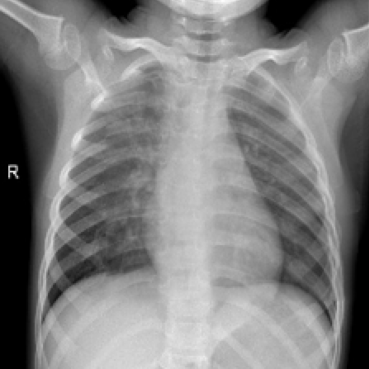}\vspace{0pt}
\includegraphics[width=1.0\linewidth]{NORMAL-1012843-0001_fpi_mask.png}\vspace{0pt}
\includegraphics[width=1.0\linewidth]{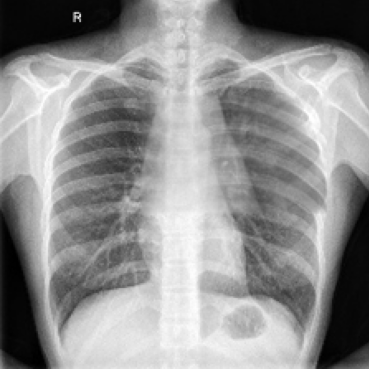}\vspace{0pt}
\includegraphics[width=1.0\linewidth]{NORMAL-9196587-0001_fpi_mask.png}\vspace{0pt}
\end{minipage} \hspace{-8pt}
}
\subfloat[CutPaste]{
\begin{minipage}[b]{0.16\linewidth}
\includegraphics[width=1.0\linewidth]{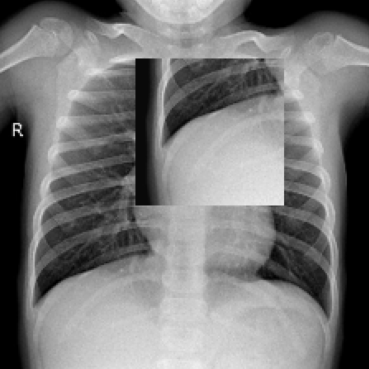}\vspace{0pt}
\includegraphics[width=1.0\linewidth]{NORMAL-28501-0001_fpi_mask.png}\vspace{0pt}
\includegraphics[width=1.0\linewidth]{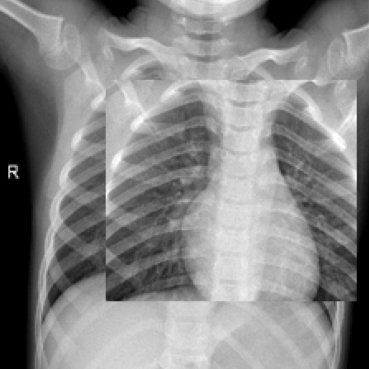}\vspace{0pt}
\includegraphics[width=1.0\linewidth]{NORMAL-1012843-0001_fpi_mask.png}\vspace{0pt}
\includegraphics[width=1.0\linewidth]{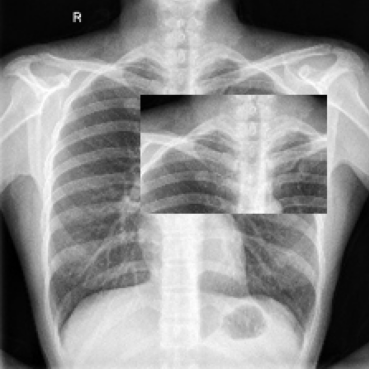}\vspace{0pt}
\includegraphics[width=1.0\linewidth]{NORMAL-9196587-0001_fpi_mask.png}\vspace{0pt}
\end{minipage} \hspace{-8pt}
}
\subfloat[AnatPaste]{
\begin{minipage}[b]{0.16\linewidth}
\includegraphics[width=1.0\linewidth]{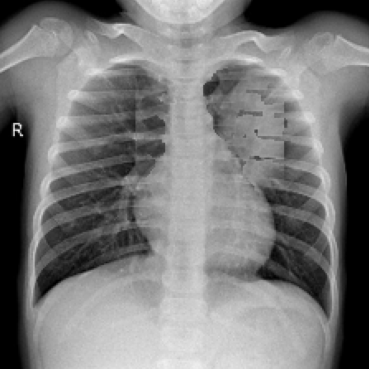}\vspace{0pt}
\includegraphics[width=1.0\linewidth]{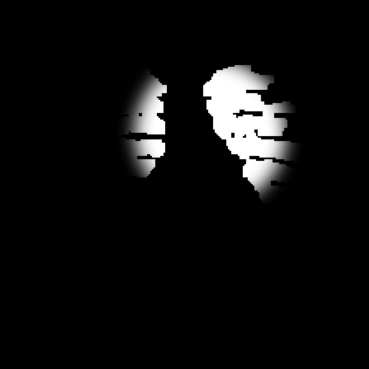}\vspace{0pt}
\includegraphics[width=1.0\linewidth]{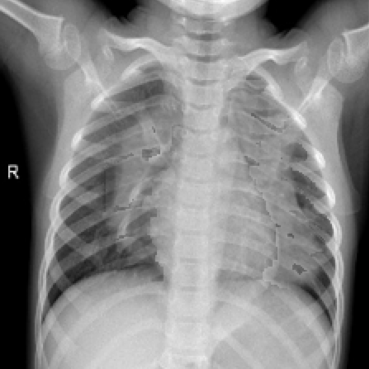}\vspace{0pt}
\includegraphics[width=1.0\linewidth]{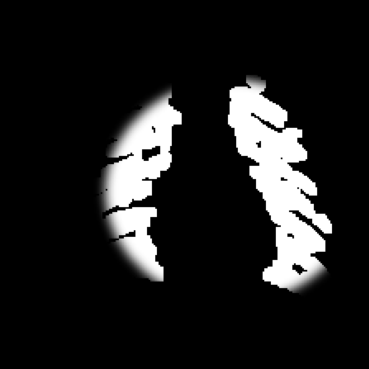}\vspace{0pt}
\includegraphics[width=1.0\linewidth]{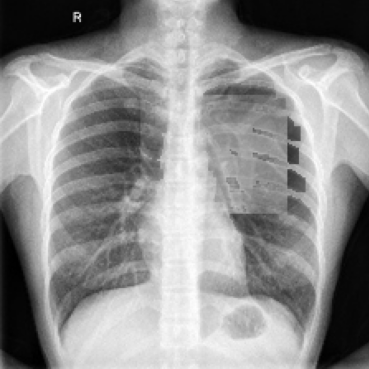}\vspace{0pt}
\includegraphics[width=1.0\linewidth]{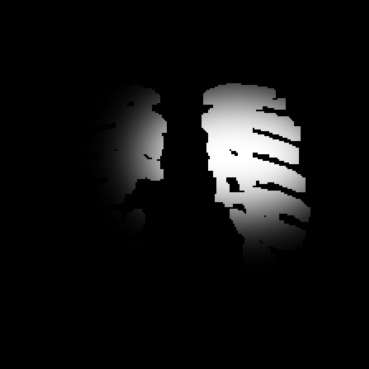}\vspace{0pt}
\end{minipage} \hspace{-8pt}
}
\subfloat[SAS]{
\begin{minipage}[b]{0.16\linewidth}
\includegraphics[width=1.0\linewidth]{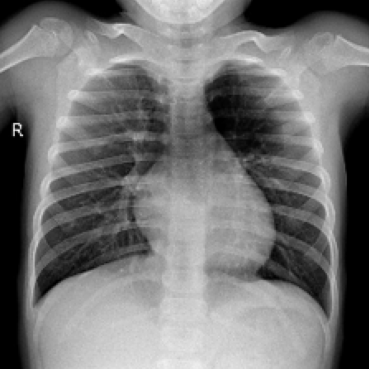}\vspace{0pt}
\includegraphics[width=1.0\linewidth]{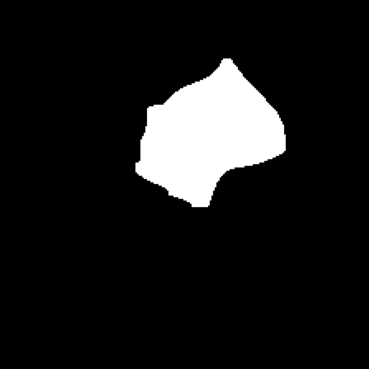}\vspace{0pt}
\includegraphics[width=1.0\linewidth]{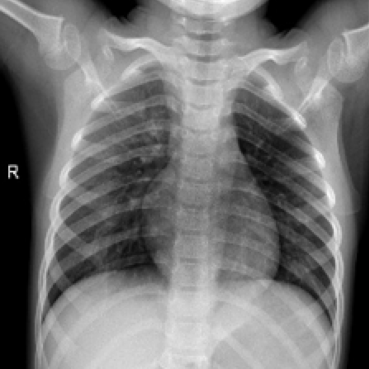}
\includegraphics[width=1.0\linewidth]{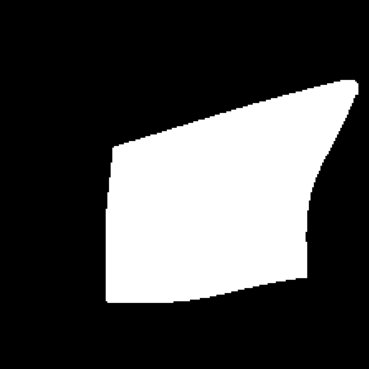}\vspace{0pt}
\includegraphics[width=1.0\linewidth]{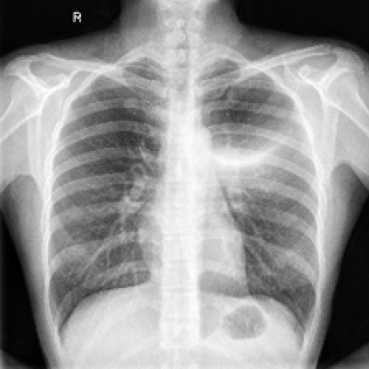}\vspace{0pt}
\includegraphics[width=1.0\linewidth]{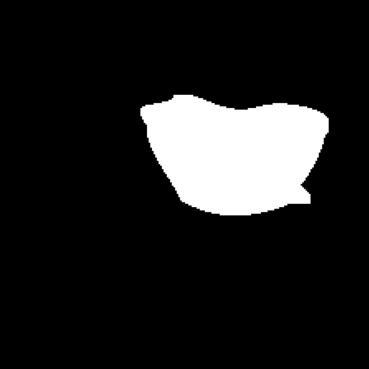}\vspace{0pt}
\end{minipage} \hspace{-8pt}
}
\caption{Visual comparison of anomaly synthesis methods on ZhangLab dataset. For each group, the first row represents visualization performance, while the second row shows the masks of the corresponding anomalies. The proposed SAS is evidenced to generate authentic anomalies while 
preserving the structure of the lung.}

\label{fig:vis}

\end{figure}
\newpage

\bibliographystyle{splncs04}
\bibliography{Paper-0578}

\begin{thebibliography}{10}
\providecommand{\url}[1]{\texttt{#1}}
\providecommand{\urlprefix}{URL }
\providecommand{\doi}[1]{https://doi.org/#1}

\bibitem{baugh2023many}
Baugh, M., Tan, J., M{\"u}ller, J.P., Dombrowski, M., Batten, J., Kainz, B.: Many tasks make light work: Learning to localise medical anomalies from multiple synthetic tasks. In: Proc. Int. Conf. Med. Image Comput. Comput.-Assisted Intervention. pp. 162--172 (2023)

\bibitem{amae}
Bozorgtabar, B., Mahapatra, D., Thiran, J.P.: {AMAE}: Adaptation of pre-trained masked autoencoder for dual-distribution anomaly detection in chest x-rays. In: Proc. Int. Conf. Med. Image Comput. Comput.-Assisted Intervention. pp. 195--205 (2023)

\bibitem{ddad}
Cai, Y., Chen, H., Yang, X., Zhou, Y., Cheng, K.T.: Dual-distribution discrepancy with self-supervised refinement for anomaly detection in medical images. Medical Image Analysis  \textbf{86},  102794 (2023)

\bibitem{chen2023knowledge}
Chen, X., He, Y., Xue, C., Ge, R., Li, S., Yang, G.: Knowledge boosting: Rethinking medical contrastive vision-language pre-training. In: Proc. Int. Conf. Med. Image Comput. Comput.-Assisted Intervention. pp. 405--415 (2023)

\bibitem{chen2022deep}
Chen, Y., Tian, Y., Pang, G., Carneiro, G.: Deep one-class classification via interpolated gaussian descriptor. In: Proc. AAAI Conf. Artif. Intell. vol.~36, pp. 383--392 (2022)

\bibitem{gong2019memorizing}
Gong, D., Liu, L., et~al.: Memorizing normality to detect anomaly: Memory-augmented deep autoencoder for unsupervised anomaly detection. In: Int. Conf. Comput. Vis. pp. 1705--1714 (2019)

\bibitem{chexpert}
Irvin, J., Rajpurkar, P., Ko, M., Yu, Y., Ciurea-Ilcus, S., Chute, C., Marklund, H., Haghgoo, B., Ball, R., Shpanskaya, K., et~al.: {CheXpert}: A large chest radiograph dataset with uncertainty labels and expert comparison. In: Proc. AAAI Conf. Artif. Intell. vol.~33, pp. 590--597 (2019)

\bibitem{johnson2019mimic}
Johnson, A.E., Pollard, T.J., Berkowitz, S.J., Greenbaum, N.R., Lungren, M.P., Deng, C.y., Mark, R.G., Horng, S.: {MIMIC-CXR}, a de-identified publicly available database of chest radiographs with free-text reports. Scientific data  \textbf{6}(1), ~317 (2019)

\bibitem{zhanglab}
Kermany, D.S., Goldbaum, M., Cai, W., Valentim, C.C., Liang, H., Baxter, S.L., McKeown, A., Yang, G., Wu, X., Yan, F., et~al.: Identifying medical diagnoses and treatable diseases by image-based deep learning. cell  \textbf{172}(5),  1122--1131 (2018)

\bibitem{khattak2023maple}
Khattak, M.U., Rasheed, H., et~al.: {MaPLe}: Multi-modal prompt learning. In: IEEE Conf. Comput. Vis. Pattern Recog. pp. 19113--19122 (2023)

\bibitem{li2021cutpaste}
Li, C.L., Sohn, K., Yoon, J., Pfister, T.: {CutPaste}: Self-supervised learning for anomaly detection and localization. In: IEEE Conf. Comput. Vis. Pattern Recog. pp. 9664--9674 (2021)

\bibitem{if2d}
Naval~Marimont, S., Tarroni, G.: Implicit field learning for unsupervised anomaly detection in medical images. In: Proc. Int. Conf. Med. Image Comput. Comput.-Assisted Intervention. pp. 189--198 (2021)

\bibitem{vindr}
Nguyen, H.Q., Lam, K., Le, L.T., Pham, H.H., Tran, D.Q., Nguyen, D.B., Le, D.D., Pham, C.M., Tong, H.T., Dinh, D.H., et~al.: {VinDr-CXR}: An open dataset of chest x-rays with radiologist’s annotations. Scientific Data  \textbf{9}(1), ~429 (2022)

\bibitem{xplainer}
Pellegrini, C., Keicher, M., {\"O}zsoy, E., Jiraskova, P., Braren, R., Navab, N.: Xplainer: From x-ray observations to explainable zero-shot diagnosis. In: Proc. Int. Conf. Med. Image Comput. Comput.-Assisted Intervention. pp. 420--429 (2023)

\bibitem{clip}
Radford, A., Kim, J.W., Hallacy, C., Ramesh, A., Goh, G., Agarwal, S., Sastry, G., Askell, A., Mishkin, P., Clark, J., et~al.: Learning transferable visual models from natural language supervision. In: Proc. Int. Conf. Mach. Learn. pp. 8748--8763 (2021)

\bibitem{reiss2021panda}
Reiss, T., Cohen, N., Bergman, L., Hoshen, Y.: {PANDA}: Adapting pretrained features for anomaly detection and segmentation. In: IEEE Conf. Comput. Vis. Pattern Recog. pp. 2806--2814 (2021)

\bibitem{mkd}
Salehi, M., Sadjadi, N., Baselizadeh, S., Rohban, M.H., Rabiee, H.R.: Multiresolution knowledge distillation for anomaly detection. In: IEEE Conf. Comput. Vis. Pattern Recog. pp. 14902--14912 (2021)

\bibitem{sato2023anatomy}
Sato, J., Suzuki, Y., Wataya, T., Nishigaki, D., Kita, K., Yamagata, K., Tomiyama, N., Kido, S.: Anatomy-aware self-supervised learning for anomaly detection in chest radiographs. iScience  (2023)

\bibitem{fpi}
Tan, J., Hou, B., Batten, J., Qiu, H., Kainz, B., et~al.: Detecting outliers with foreign patch interpolation. Machine Learning for Biomedical Imaging  \textbf{1},  1--27 (2022)

\bibitem{pii}
Tan, J., Hou, B., Day, T., Simpson, J., Rueckert, D., Kainz, B.: Detecting outliers with poisson image interpolation. In: Proc. Int. Conf. Med. Image Comput. Comput.-Assisted Intervention. pp. 581--591 (2021)

\bibitem{chexzero}
Tiu, E., Talius, E., Patel, P., Langlotz, C.P., Ng, A.Y., Rajpurkar, P.: Expert-level detection of pathologies from unannotated chest x-ray images via self-supervised learning. Nature Biomedical Engineering  \textbf{6}(12),  1399--1406 (2022)

\bibitem{squid}
Xiang, T., Zhang, Y., Lu, Y., Yuille, A.L., Zhang, C., Cai, W., Zhou, Z.: {SQUID}: Deep feature in-painting for unsupervised anomaly detection. In: IEEE Conf. Comput. Vis. Pattern Recog. pp. 23890--23901 (2023)

\bibitem{you2023cxr}
You, K., Gu, J., Ham, J., Park, B., Kim, J., Hong, E.K., Baek, W., Roh, B.: {CXR-CLIP}: Toward large scale chest x-ray language-image pre-training. In: Proc. Int. Conf. Med. Image Comput. Comput.-Assisted Intervention. pp. 101--111 (2023)

\bibitem{salad}
Zhao, H., Li, Y., He, N., Ma, K., Fang, L., Li, H., Zheng, Y.: Anomaly detection for medical images using self-supervised and translation-consistent features. IEEE Trans. Medical Imaging.  \textbf{40}(12),  3641--3651 (2021)

\bibitem{zhou2022learning}
Zhou, K., Yang, J., Loy, C.C., Liu, Z.: Learning to prompt for vision-language models. Int. J. Comput. Vis.  \textbf{130}(9),  2337--2348 (2022)

\bibitem{zhou2021proxy}
Zhou, K., Li, J., Luo, W., Li, Z., Yang, J., Fu, H., Cheng, J., Liu, J., Gao, S.: Proxy-bridged image reconstruction network for anomaly detection in medical images. IEEE Trans. Medical Imaging.  \textbf{41}(3),  582--594 (2021)

\bibitem{zhou2020encoding}
Zhou, K., Xiao, Y., Yang, J., Cheng, J., Liu, W., Luo, W., Gu, Z., Liu, J., Gao, S.: Encoding structure-texture relation with p-net for anomaly detection in retinal images. In: Eur. Conf. Comput. Vis. pp. 360--377 (2020)

\end{thebibliography}

\end{document}